\documentclass{Interspeech}

\interspeechcameraready

\usepackage{cite}
\usepackage{amsmath,amssymb,amsfonts}
\usepackage{algorithmic}
\usepackage{graphicx}
\usepackage{textcomp}
\usepackage{xcolor}

\usepackage{color, tabularx}
\usepackage{booktabs, multirow}
\usepackage{float}
\usepackage{url}
\usepackage{color, colortbl}
\definecolor{Gray}{gray}{0.9}
\usepackage{diagbox}
\usepackage{hyperref}

\usepackage{pifont}
\newcommand{\cmark}{\ding{51}}%
\newcommand{\xmark}{\ding{55}}%

\title{Enhancing Low-Resource Language and Instruction Following Capabilities of Audio Language Models}

\author[affiliation={1}]{Potsawee}{Manakul}
\author[affiliation={2}]{Guangzhi}{Sun}
\author[affiliation={1}]{Warit}{Sirichotedumrong}
\author[affiliation={1}]{Kasima}{Tharnpipitchai}
\author[affiliation={1}]{Kunat}{Pipatanakul}

\affiliation{SCB 10X}{SCBX Group}{Thailand}
\affiliation{Department of Engineering}{University of Cambridge}{United Kingdom}
\email{\{potsawee, warit, kasima, kunat\}@scb10x.com, gs534@cam.ac.uk}


\usepackage{comment}

\begin{document}

\maketitle

\begin{abstract}
    
    Audio language models process audio inputs using textual prompts for tasks like speech recognition and audio captioning. Although built on multilingual pre-trained components, most are trained primarily on English, limiting their usability for other languages. This paper evaluates audio language models on Thai, a low-resource language, and finds that they lack emergent cross-lingual abilities despite their multilingual foundations. To address this, we explore data mixtures that optimize audio language models for both a target language and English while integrating audio comprehension and speech instruction-following into a unified model. Our experiments provide insights into improving instruction-following in low-resource languages by balancing language-specific and multilingual training data. The proposed model, Typhoon-Audio, significantly outperforms existing open-source models and achieves performance comparable to state-of-the-art Gemini-1.5-Pro in both English and Thai.
\end{abstract}

\section{Introduction}
Recent advancements in multimodal large language models (LLMs) have involved augmenting audio encoders, allowing models to extend their capabilities beyond textual inputs to include the understanding of speech and audio events \cite{chu2023qwen, tang2024salmonn, gong2024listen, gong2023ltuas, zhang2023speechgpt, fathullah-etal-2024-audiochatllama, hu2024wavllm}. Audio language models are developed for different use cases with common types as follows. Audio-understanding models, e.g., Qwen-Audio \cite{chu2023qwen}, SALMONN \cite{tang2024salmonn}, and LTU \cite{gong2024listen}, can process various types of audio inputs, such as speech, environmental sounds, or music, and generate textual outputs. These models are primarily designed for tasks that require a deep understanding of audio content, prompted by human instructions such as speech recognition, speech translation, or audio captioning. Alternatively, models such as SpeechGPT~\cite{zhang2023speechgpt} accept speech as input and produce speech as output, functioning as voice agents. These models are optimized for seamless voice interactions rather than complex audio understanding. This work focuses on audio-understanding and instruction-following models, which we refer to as audio language models. 

Audio language models typically comprise three key components: an audio encoder backbone, an LLM backbone, and an adapter module, as outlined in Section~\ref{section:related_work}. Despite leveraging multilingual backbones, most models are primarily trained on: (1) English data, and (2) only audio content understanding tasks, as seen in models like Qwen-Audio~\cite{chu2023qwen} and SALMONN~\cite{tang2024salmonn}, or only speech instruction understanding, such as AudioChatLlama~\cite{fathullah-etal-2024-audiochatllama}. Addressing these limitations, this work focuses on two goals. {First}, we examine model performance in a low-resource language using Thai as a case study, and we provide a recipe to enhance the low-resource language ability while retaining the English performance. {Second}, we integrate \textit{improved} audio-understanding and speech instruction understanding capabilities into one unified model. 

\section{Related Work}
\label{section:related_work}
\noindent \textbf{Audio Language Models}: 
\textit{SALMONN} integrates three primary components: an LLM based on Vicuna \cite{vicuna2023}, a speech encoder based on Whisper-large-v2 \cite{radford2023robust}, and BEATS \cite {chen2023beats} for audio events. The representations from Whisper and BEATS are concatenated and passed through an adapter (Q-Former) to obtain the audio representation as the input to the LLM. Trainable modules are the Q-Former and the LoRA weights of the LLM. Training data consists of speech recognition, translation, audio captioning, or spoken QA. 
\textit{Qwen-Audio}, similarly, uses Whisper-large-v2 encoder, and its LLM is based on Qwen \cite{bai2023qwen}. No connection module (adapter) is employed; however, a variety of special tokens are incorporated. The pre-training phase involves training the audio encoder while freezing the LLM. During the fine-tuning phase, the audio encoder is frozen, and only the LLM is trained.
\textit{LTU} (first version) \cite{gong2024listen} was developed to support audio events with Llama-7B (LLM) and an AST audio encoder \cite{gong21b_interspeech} as backbones. The AST encoder maps audio into 32 tokens, with a connection module adjusting dimensions from 768 to 4096. Subsequently, \textit{LTU-AS} (second version) \cite{gong2023ltuas} extends support to both speech and audio, using Whisper-encoder, similar to SALMONN and Qwen-Audio. The training involved a curated dataset, with around 5 million triples (audio, question, answer) in the first version, and expanded to 9 million triples in the second version. These training examples were generated using GPT-3.5 based on audio metadata. Alternatively, \textit{AudioChatLlama} aligns the semantic space of speech and text for direct speech-to-response generation. Since AudioChatLlama is optimized for responding to speech input, it cannot be controlled to perform other tasks through text instructions like SALMONN, Qwen-Audio, or LTU.

\vspace{1mm}
\noindent \textbf{LLMs for Low-Resource Language}: Previous studies have adapted English-centric unimodal LLMs into bilingual unimodal LLMs, enabling them to perform effectively in both English and a target language \cite{zeng2023glmb, pipatanakul2023typhoon, cui2023efficient, csaki2024sambalingo}. For example, Typhoon \cite{pipatanakul2023typhoon} is a Mistral/Llama3-based LLM enhanced on Thai. Typhoon was continually pre-trained on 50\% English and 50\% Thai data where Thai texts were sourced from the MC4 and OSCAR datasets before supervised fine-tuning. This process showed an improvement in Thai evaluation benchmarks. We note that while adapting LLMs to a new language is well-explored in unimodal LLMs, research on adapting audio language models to a new language remains underexplored.

\section{Methodology}
\begin{figure}[!ht]
    \centerline{
\includegraphics[width=\linewidth,keepaspectratio]{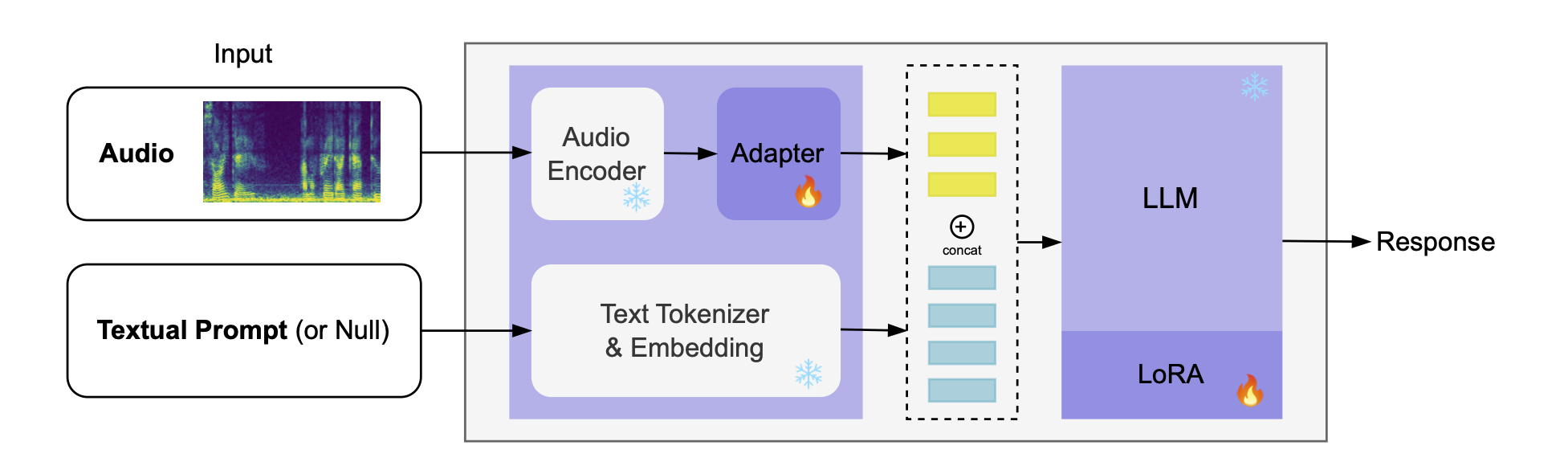}}
    \caption{The model architecture of Typhoon-Audio. The audio encoder consists of a Whisper encoder and a BEATs encoder. The adapter is based on a window-level Q-Former. The LLM is our Typhoon model.}
    \label{fig:architecture}
\end{figure}

\noindent \textbf{Model Architecture} (in Fig.~\ref{fig:architecture}): We follow the SALMONN architecture~\cite{tang2024salmonn}. With Thai and English as target languages, our model is based on Typhoon-1.5–8B-Instruct~\cite{pipatanakul2023typhoon} as the LLM, Whisper-large-v3 fine-tuned to Thai \cite{thonburian_whisper_med} coupled with BEATs~\cite{chen2023beats} as the audio encoder, and Q-Former~\cite{li2023blip2} trained from scratch as the adapter. Note that we examine other variants of LLM and audio encoder backbones in Section~\ref{section:exp_pretraining}.

\label{sec:architecture}

\noindent  \textbf{Training Strategies}: The audio encoder maps the spectrogram into a representation, which is then transformed into the audio representation $a$ in the text embedding space via an adapter. The model $\theta$ is trained to maximize the probability of the next word $y_t$ in the textual response, conditioned on previous words $y_{1:t-1}$, textual prompt input $x$ and the audio input $a$: $P(y_t | y_{1:t-1}, x, a; \theta)$. Training occurs in two phases:

\vspace{1mm}
\scalebox{0.9}{$\bullet$} \textit{Pre-training}: As the adapter is the only component initialized with random weights, this phase trains only the adapter to align audio and textual representations. We use ASR and audio captioning data shown in Tab.~\ref{tab:training_data} in this phase. 

\vspace{1mm}
\scalebox{0.9}{$\bullet$} \textit{Supervised Fine-Tuning (SFT)}: This phase trains both the adapter and the LoRA weight \cite{hu2022lora} of the LLM ($r$=8, $\alpha$=32). During SFT, the model is trained on diverse tasks and instruction prompts to enhance its instruction-following capabilities. Tab.~\ref{tab:sft_data} presents the data used in our final model (Typhoon-Audio), and Section~\ref{section:sft} studies the SFT data mixture.

\section{Experimental Setup}
\label{section:tasks}

\begin{table}[!ht]
    \small
    \tabcolsep=0.6mm
    \caption{Pre-training data -- 1.82M examples in total}
    \centering
    \begin{tabular}{lccc}
    \toprule
        Dataset & Task  &Lang & \#Examples  \\ 
        \midrule
        LibriSpeech \cite{panayotov2015librispeech}   &ASR          &En  &281K  \\
        GigaSpeech-M \cite{gigaSpeech2021}   &ASR          &En  &900K  \\
        CommonVoice-Th \cite{commonvoice2020} &ASR          &Th  &436K  \\
        Fleurs-Th \cite{fleurs2022arxiv}      &ASR          &Th  &7.8K  \\
        Vulcan+Elderly+Gowajee         &ASR          &Th  &65.1K  \\
        \midrule
        AudioCaps \cite{audiocaps}           &AudioCaption &En+Th  &48.3+48.3K  \\
        Clotho \cite{clotho}             &AudioCaption &En+Th  &19.2+19.2K  \\
    \bottomrule
    \end{tabular}
    \label{tab:training_data}
\end{table}

\begin{table}[!ht]
    \footnotesize
    \tabcolsep=1mm
    \caption{SFT data of Typhoon-Audio -- 640K examples in total}
    \centering
    \begin{tabular}{llcc}
    \toprule
        Dataset &Task &New &\#Ex  \\ 
        \midrule
        \rowcolor{Gray}
        \multicolumn{4}{l}{QA pairs from SALMONN used in \texttt{SFT-v1, SFT-v2, SFT-v3}}  \\
        LibriSpeech \cite{panayotov2015librispeech}    &QA (Speech-En) &\xmark   &40.0K  \\
        AudioCaps \cite{audiocaps}     &QA (Audio)  &\xmark  &30.0K  \\
        \rowcolor{Gray}
        \multicolumn{4}{l}{QA pairs from LTU-AS used in \texttt{SFT-v1, SFT-v2, SFT-v3}}  \\
        LibriTTS \cite{libritts}          &QA (Speech-En)  &\xmark &21.1K  \\
        IEMOCAP \cite{busso2008iemocap}           &QA (Speech-En)  &\xmark &4.3K   \\
        FSD50K  \cite{fonseca2021fsd50k}           &QA (Audio)   &\xmark &11.5K  \\
        AudioSet  \cite{audioset}         &QA (Audio-Speech) &\xmark & 20.0K \\
        AS20k  \cite{asstrong}            &QA (Audio-Speech) &\xmark & 12.0K \\
        \rowcolor{Gray}
        \multicolumn{4}{l}{ASR, Translation, Audio Caption, QA used in \texttt{SFT-v2, SFT-v3}} \\
        LibriSpeech \cite{panayotov2015librispeech}   &ASR (En)         &\xmark &32.0K \\
        CommonVoice-Th \cite{commonvoice2020}   &ASR (Th)         &\xmark &52.0K \\
        SelfInstruct-Th &ASR (Th)         &\cmark &18.9K  \\
        AudioCaps(Gemini) & Audio Caption &\cmark &48.3K \\
        Covost2 \cite{covost2}       &Translate (X2Th) &\xmark &30.0K \\
        CommonVoice-Th \cite{commonvoice2020}   &Translate (Th2X) &\xmark &7.3K  \\
        VISTEC-SER \cite{vistec_ser}   &QA (Emotion\&Gender) &\cmark    &18.0K \\
        Yodas2-30S \cite{li2023yodas}  &QA (Speech-Th)  &\cmark    &90.0K \\ 
        \rowcolor{Gray}
        \multicolumn{4}{l}{Speech Instruction Following used in \texttt{SFT-v3}} \\
        GigaSpeech \cite{gigaSpeech2021}         &SpeechIF-Type1 (En) &\cmark &20.0K \\
        CommonVoice-Th \cite{commonvoice2020}   &SpeechIF-Type1 (Th) &\cmark    &120.5K  \\
        jan-hq-instruction-v1 \cite{janhq} &SpeechIF-Type2 (En) &\xmark &20.0K \\
        Airoboros-Th  &SpeechIF-Type2 (Th)  &\cmark    &5.7K   \\
        Alpaca-Th     &SpeechIF-Type2 (Th) &\cmark    &20.0K  \\
        SelfInstruct-Th &SpeechIF-Type2 (Th) &\cmark    &18.9K  \\
    \bottomrule
    \end{tabular}
    \label{tab:sft_data}
\end{table}

\noindent \textbf{Data}: Each example has \{audio, textual\_prompt\}. For \textit{pre-training} data (Tab.~\ref{tab:training_data}), a few prompts were pre-defined for each task, e.g., ``Transcribe this audio" for ASR, and we sample a prompt language that matches the response language. Since there is no Thai audio-captioning data, we translate AudioCaps and Clotho to Thai. For \textit{SFT} data (Tab.~\ref{tab:sft_data}), we translate 10\% of prompts/responses in existing QA data to Thai. To increase the diversity of prompts, we generate a prompt for each example in ASR, translation, and audio-captioning, using GPT-4o. For speech instruction following, the model is meant to listen to spoken instructions and respond, thus, the prompt is null. Next, we describe each SFT task in more detail as follows:


\scalebox{0.9}{$\bullet$} \textit{ASR}: Datasets are used as shown in Table~\ref{tab:sft_data}. 
A prompt is, for example, ``Transcribe this audio" as well as its paraphrases.

\scalebox{0.9}{$\bullet$} \textit{Audio Caption}: This task involves generating audio descriptions, and we use AudioCaps \cite{audiocaps}, with English references translated into Thai for Thai Audio Captioning. Although AudioCaps was used for pre-training, its short ground-truth captions limit detailed response generation during SFT. To address this limitation, we provide Gemini-1.5-Pro with both audio input and the short caption, prompting it to generate detailed responses. This augmented data is called AudioCaps (Gemini).

\scalebox{0.9}{$\bullet$} \textit{Speech Translation}: Thai-to-English is from CommonVoice (Thai), and target English texts are derived from translation. English/X-to-Thai is from Covost2, and target Thai texts are derived from translating English texts. X refers to a non-English audio language (Arabic, German, Spanish, French, Indonesian, Italian, Japanese, Chinese) taken from Covost2. The translation was performed using our internal system, which matches Google Translate API performance. Each setup includes 2000 examples. The evaluation metric is BLEU.


\scalebox{0.9}{$\bullet$} \textit{Spoken Document QA}: The QA examples in SALMONN and LTU are primarily based on short utterances, typically under 10 seconds. To achieve longer understanding capabilities, we chunked longer audio in Yodas2~\cite{li2023yodas} into 30-second segments and prompted GPT-4o to generate question-answer pairs from the reference texts. In addition to standard extractive QA pairs, we also generated multiple-choice questions (MCQ), as we found that this approach improved the SpokenQA performance, which will be discussed in Section~\ref{section:sft}. Furthermore, we focus on the Thai subset of Yodas2, as existing QA datasets predominantly consist of English spoken documents. Since Yodas2 emphasizes the content of speech, to capture the unique characteristics of voices, we additionally generated QA pairs from the VISTEC-SER dataset~ \cite{vistec_ser}, which includes metadata such as speaker gender and emotion. The evaluation metric is word overlap of generated response against  reference using F1.




\begin{figure}[!h]
    \centerline{
\includegraphics[width=\linewidth,keepaspectratio]{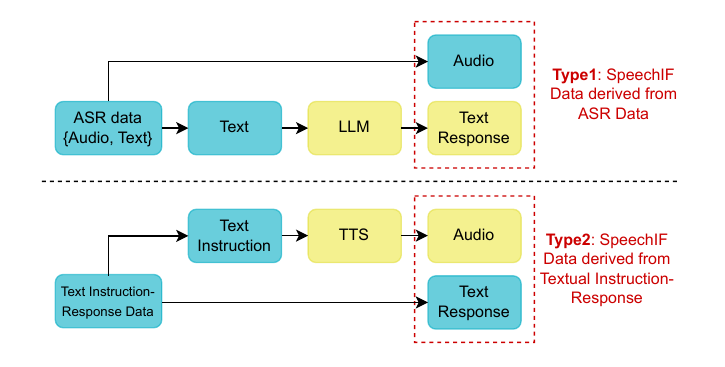}}
    \caption{Speech Instruction Following Data Creation Pipeline}
    \label{fig:speech_if}
\end{figure}

\scalebox{0.9}{$\bullet$} \textit{Speech Instruction Following (SpeechIF)}: This task requires models to listen to spoken instructions and directly respond. Current models like SALMONN lack specific data for this ability. We propose two methods for generating SpeechIF data (Fig.~\ref{fig:speech_if}). \textit{Type1} leverages ASR datasets to generate text responses from transcripts. However, since ASR data typically contains non-question utterances, LLMs often default to safe responses such as ``I'm sorry, as an AI assistant I cannot..." in up to 30\% of cases. While it offers voice diversity, it does not fully reflect real-world interactions. \textit{Type2} synthesizes speech from instruction-response pairs (e.g., Alpaca, Airoboros), providing more practical commands but struggles with unsuitable instructions like math or coding. Though lacking voice diversity, Type2 represents real interactions better. This work utilizes both types as shown in Table~\ref{tab:sft_data}. For evaluation, we selected instructions from AlpacaEval (English) and SelfInstruct (Thai), creating a  SpeechIF benchmark for both languages. The prompt for evaluating baseline models (e.g., SALMONN) is ``Listen to the audio and answer the question". 

\scalebox{0.9}{$\bullet$} \textit{Complex Instruction Following (ComplexIF)}: We propose ComplexIF to assess models' ability to follow unseen, compound instructions, where each instruction involves two to three audio tasks (such as transcribe, then translate). In ComplexIF, models have to respond in specific formats (e.g., JSON, XML), with format templates provided in the instruction prompt. As it evaluates the general instruction following ability, only English speech data is used. ComplexIF is used exclusively for evaluation, without additional training.

\vspace{2mm}
\noindent \textbf{Evaluation}: For existing tasks, we use standard metrics. For SpeechIF and ComplexIF, we follow the setup of MT-Bench in using an LLM judge (GPT-4o), and we adapt the single-turn evaluation prompt from MT-Bench and score responses on a scale from 1.0 to 10.0. For ComplexIF, we prompt the judge to evaluate the response on two aspects: 

\begin{itemize}
\item \textit{Quality} considers helpfulness, relevance, accuracy, depth, creativity, and level of detail of the response.

\item \textit{Format} considers how well the response follows the format required by the user (e.g., JSON, XML, Markdown, etc).
\end{itemize}
\vspace{2mm}
\noindent \textbf{Baselines}: Competitive audio language models include,
\begin{itemize}
    \item \textit{Open-weights}: Qwen-Audio (Qwen-7B) \cite{chu2023qwen}, SALMONN (Vicuna-13B) \cite{tang2024salmonn}, and DiVA (Llama3-8B) \cite{held2024diva}.
    \item \textit{Proprietary}: Gemini-1.5-Pro (Audio)
\end{itemize}
For open models, we use available weights on HuggingFace. For Gemini-1.5-Pro, we use \texttt{gemini-1.5-pro-001} through Google API with \{audio, text\_instruction\} as input.

\section{Results and Key Findings}
\subsection{Existing Audio Language Models on English vs Thai}
The results in Table~\ref{tab:main_results} demonstrate that: (1) Baseline models (Qwen-Audio, SALMONN, DiVA) despite using multilingual backbones exhibit significant performance degradation in Thai. Gemini-1.5-Pro maintains strong performance across both Thai and English. (2) Among the baselines, DiVA is the only model that performs well on the SpeechIF task, but it experiences a notable drop when tested on Thai. Thus, the subsequent experiments aim to develop a model that can effectively handle these tasks in both English and a low-resource language such as Thai.


\subsection{Pre-training: Speech Encoder and LLM Backbones}
\label{section:exp_pretraining}
This experiment focuses on selecting backbones, and comparing Whisper with its English+Thai fine-tuned variant. Similarly, Typhoon is a Llama-3 model fine-tuned to English+Thai. Our results (in Tab.~\ref{tab:pretraining_results}) show that for ASR, models where both backbones are matched with the target language yield the best results. However, for audio captioning, the performance difference between these models is marginal. As a low-resource language such as Thai is our goal, Whisper-Th coupled with Typhoon-1.5 are selected for subsequent experiments.

\begin{table}[!h]
    \tabcolsep=1.1mm
    \caption{Pre-training Results on ASR: LibriSpeech (other), CommonVoice (*subset-1K), AC: AudioCaps (En \& Th). Whisper = Whisper-v3-large.}
    \centering
    \begin{tabular}{ll|cccc}
    \toprule
        \multirow{2}{*}{SpeechEnc} &\multirow{2}{*}{LLM}  &\multicolumn{2}{c}{ASR (WER$\downarrow$)} &\multicolumn{2}{c}{AC (METEOR$\uparrow$)} \\
        &&En &Th* &En &Th  \\ 
        \midrule
        Whisper        &Llama-3      &\textbf{6.02} &16.66 &30.75 &20.04  \\
        Whisper        &Typhoon-1.5           &7.76 &20.01 &29.56 &\textbf{20.62}  \\
        Whisper-Th     &Llama-3      &7.35 &15.68 &29.52 &19.94  \\
        Whisper-Th     &Typhoon-1.5           &9.15 &\textbf{13.52} &\textbf{30.83} &20.55  \\
    \bottomrule
    \end{tabular}
    \label{tab:pretraining_results}
\end{table}

\begin{table*}[!t]
    \tabcolsep=1mm
    \caption{Audio LM Evaluation in English and Thai. ASR: En=LibriSpeech-other, Th=CommonVoice-17; Translation: Th-to-En=CommonVoice-17, En/X-to-Th=Covost2 where reference texts are derived from translation; Gender Classification: En \& Th = Fleurs; SpokenQA: En=SpokenSQuAD~\cite{lee2018spoken}, Th=CommonVoice-17 where QA are generated from references using GPT-4o; SpeechIF: En=AlpcaEval-TTS, Th=SelfInstruct-TTS. ComplexIF: Mixture of 5 other tasks in English. SpeechIF/ComplexIF $\in$ [1,10]. $^*$Typhoon2-Audio follows the same training methodologies of Typhoon-Audio, except for the change in the base LLM from Typhoon1.5 to Typhoon2. $^\dagger$We observed examples where the model did not provide any answer to nested commands; hence, receiving low scores.}
    \centering
    \begin{tabular}{cc|cc|ccc|cc|cc|cc|cc}
    \toprule
        \multirow{2}{*}{Model} &\multirow{2}{*}{Size} &\multicolumn{2}{c}{ASR (WER$\downarrow$)} &\multicolumn{3}{c}{Translation (BLEU$\uparrow$)}  &\multicolumn{2}{c}{Gender (Acc$\uparrow$)}  &\multicolumn{2}{c}{SpQA (F1$\uparrow$)} &\multicolumn{2}{c}{SpeechIF (Judge$\uparrow$)}  &\multicolumn{2}{c}{CpxIF (Judge$\uparrow$)}  \\ 
        & &En &Th &Th2En &En2Th &X2Th &En &Th &En &Th &En &Th &Qual &Format \\ 
        \midrule
        Qwen-Audio \cite{chu2023qwen}    &7B     &6.94  &95.12 &0.00 &2.48 &0.29 &37.09 &67.97  &25.34 &0.00 &1.07 &1.03 &3.13 &1.68  \\ 
        SALMONN \cite{tang2024salmonn}   &13B    &\textbf{5.79}  &98.07 &14.97 &0.07  &0.10  &95.69 &93.26 &52.92 &2.95  &2.47 &1.18  &4.10 &5.09 \\ 
        DiVA \cite{held2024diva}         &8B     &30.28 &65.21 &7.97  &9.82  &5.31  &47.30 &50.12 &44.52&15.13 &\textbf{6.81} &2.68 &6.33 &7.83  \\ 
        Gemini-1.5-Pro &-  &5.98  &\textbf{13.56} &22.54 &{20.69} &{13.52} &90.73 &81.32 &\textbf{74.09} &62.10 &3.24 &3.93 &\textbf{7.25} &{8.99} \\ 
        \midrule    
        Typhoon-Audio                    &8B     &8.72  &14.17 &{24.14} &17.52 &10.67 &\textbf{98.76} &\textbf{93.74} &48.83 &{64.60} &5.62 &{6.11} &6.34 &8.73 \\ 
        Typhoon2-Audio$^*$                    &8B   &5.83  &14.04  &\textbf{33.25}  &\textbf{27.15}   &\textbf{15.93}   &76.51 &75.65 &69.22 &\textbf{70.01} &6.00 &\textbf{6.79}  &5.35$^\dagger$ &\textbf{9.01}     \\
    \bottomrule
    \end{tabular}
    \label{tab:main_results}
\end{table*}

\subsection{Supervised Fine Tuning (SFT): Data Mixture}
\label{section:sft}

\begin{table}[!h]
\tabcolsep=0.5mm
    \caption{SFT Results of different SFT recipes on various Thai Tasks and on English ComplexIF.  $^*$ASR is evaluated on the subset-1K of CV17. $^\dagger$Avg. of Qual and Format.}
    \centering
    \begin{tabular}{lc|cccc|c}
    \toprule
       Experiment  &\#Ex &ASR*$\downarrow$ &Th2En &SpQA &SpIF &CxIF$^\dagger$ \\        
        \midrule 
        Pre-trained  &- &13.52 &0.00 &28.33 &1.12 &1.41 \\ 
        \rowcolor{Gray}
        \multicolumn{7}{l}{\texttt{Typhoon-Audio (SFT-v1)}} \\
        100\% En-Prompt  &600K  &80.86 &6.01  &36.88 &1.48 &6.35 \\ 
        \rowcolor{Gray}
        \multicolumn{7}{l}{\texttt{Typhoon-Audio (SFT-v2)}} \\ 
        10\% Th-Prompt &200K  &16.80 &0.00  &35.26 &3.72 &5.08 \\ 
        + QA &220K &16.93 &0.02  &46.82 &4.29 &5.33 \\ 
        + QA+Trn &240K &18.33 &21.53 &44.93 &4.25 &5.97 \\ 
        + 2*QA+Trn+MCQ  &300K &19.84 &22.04 &61.63 &4.60 &6.31 \\ 
        \rowcolor{Gray}
        \multicolumn{7}{l}{\texttt{Typhoon-Audio (SFT-v3)}} \\ 
        ScaledUp-\texttt{v2}+SpIF &620K &19.07 &23.77 &62.79 &6.32 &6.45 \\  
        + ASR (SelfInst-Th) &640K &16.89 &24.14 &64.60 &6.11 &7.54 \\ 
    \bottomrule
    \end{tabular}
    \label{tab:sft_results}
\end{table}
This experiment focuses on data mixture to enhance broad instruction-following abilities across tasks and languages. Training is initialized using the pre-trained model from the previous section. Our findings (results in Tab.~\ref{tab:sft_results}) indicate that:

\scalebox{0.9}{$\bullet$} \textit{Pre-trained} model does not exhibit task ability and it simply provides transcriptions of speech regardless of instructions. 

\scalebox{0.9}{$\bullet$} \texttt{SFT-v1}: When fine-tuned on only English prompt-response pairs (a subset of around 600K pairs in total taken from SALMONN and LTU), the model achieves better performance on new tasks, but performs poorly on Thai ASR, showing similar characteristics to SALMONN shown in Table~\ref{tab:main_results}.

\scalebox{0.9}{$\bullet$} \texttt{SFT-v2}:  This setup translates 10\% of SALMONN \& LTU QA data into Thai and uses 10\% Thai instructions for ASR, maintaining a 90/10 language ratio. Even with around 170K of mixed language QA pairs (due to a limited training budget) and 30K ASR examples, it improves SpQA and SpIF over the pre-trained model while achieving Thai ASR WER of 16.80. By adding around 20K of Thai speech QA, generated from 30-second chunks of Yodas2 and VISTEC-SER, the model achieves higher SpQA scores, and interestingly, it provides a gain in SpIF as well. Speech translation ability is only acquired when explicitly adding around 20K translation examples in training (+Trn), though with minimal impact on other tasks. Next, we further add additional spoken QA and multiple-choice (+MCQ) of around 60K addition examples, showing further gains on SpQA, SpIF, CxIF tasks. Throughout these iterations, the improvement is also reflected in the CxIF score.

\scalebox{0.9}{$\bullet$} \texttt{SFT-v3}: This configuration enhances \texttt{SFT-v2} by incorporating speech instruction following data and increasing the total number of SFT examples by approximately twice. This approach improves performance across all tasks, especially SpeechIF. Due to limited gains from scaling, we did not construct more training examples. Compared to the pre-trained model, there is a large drop in ASR performance, with the WER increasing from 13.52 to 19.07. We observed that when the speech is a command, the model may unexpectedly give a response to the command even when tasked to perform ASR. To address this issue, we incorporated the SelfInstruct data as an ASR task in addition to the SpeechIF task. This approach lowered the WER to 16.89 while preserving other capabilities.

\subsection{Typhoon-Audio vs Existing Audio Language Models}

This experiment evaluates our best model, referred to as \textit{Typhoon-Audio}, from the SFT section, benchmarking it against competitive models with results shown in Table~\ref{tab:main_results}. For ASR, although Typhoon-Audio underperforms in English, it is one of only two models (along with Gemini) that achieves a WER under 15.0 on the Thai ASR benchmark. In translation, Typhoon-Audio surpasses SALMONN and Gemini-1.5-Pro in Thai-to-English, which requires Thai comprehension and English generation. Regarding voice characteristics, SALMONN's English gender recognition transfers to Thai, with Typhoon-Audio showing comparable performance. In spoken document QA, Typhoon-Audio matches Gemini-1.5-Pro, making it the only open-source model capable of this task in Thai. For speech instruction following, while DiVA leads in English, it lacks Thai speech comprehension. Our Typhoon-Audio outperforms Gemini-1.5-Pro in both English and Thai for this task. Moreover, when presented with complex instructions, Typhoon-Audio performs the closest to Gemini-1.5-Pro in both quality and format adherence, making it the best open-source model. 

Additionally, we explore the impact of upgrading the base LLM while using the same audio-text alignment (pre-training and SFT) methodologies. All of the experiments up to now were conducted using Typhoon1.5 as the base LLM. Given the new release of an improved version of Thai LLM (Typhoon2) at this stage of development, we investigate swapping the base LLM to Typhoon2 to develop \textbf{Typhoon2-Audio}. Results in Table~\ref{tab:main_results} show improved performance of Typhoon2-Audio over Typhoon-Audio in ASR, speech translation, spoken QA, and speech instruction following. Our qualitative evaluation also shows that despite using the same audio-alignment data mixture, Typhoon2-Audio has a lower hallucination rate and reduced code-switching tendency than Typhoon-Audio, with hallucinations also lower than previous findings~\cite{sun2024crosscheckgpt}. This suggests that the base LLM’s performance is crucial to the overall effectiveness of audio language models. However, both Typhoon-Audio versions struggle with background noise, likely due to training on mostly noise-free data, highlighting the need to incorporate noise during training for better robustness.

\section{Conclusions}
This paper demonstrates the limitations of low-resource language capabilities in open-source audio language models, using Thai as an example. Through our data mixtures which combine audio content understanding and speech instruction-following, we achieve performance on par with the proprietary Gemini-1.5-Pro in a range of audio tasks in both English and Thai with only around 1.82M pre-training and 0.64M SFT examples.

\newpage
\bibliographystyle{IEEEtran}
\bibliography{mybib}

\end{document}